\pdfoutput=1

\documentclass[11pt]{article}

\usepackage[final]{acl}
\usepackage{multirow}
\usepackage{colortbl}
\usepackage{float} 

\usepackage{amssymb}
\usepackage{times}
\usepackage{placeins} 
\usepackage{afterpage}  
\usepackage{subcaption} 
\usepackage{listings}
\usepackage{amsmath}
\usepackage[most]{tcolorbox} 

\usepackage{latexsym}
\usepackage{booktabs}
\usepackage{array}
\usepackage{longtable}
\usepackage{xcolor}
\usepackage{booktabs}
\usepackage{arydshln}
\usepackage{array}
\usepackage[T1]{fontenc}

\usepackage[utf8]{inputenc}

\usepackage{microtype}

\usepackage{inconsolata}

\usepackage{graphicx}

\definecolor{lightblue}{RGB}{210, 220, 250}
\newcommand{\blue}{\cellcolor{lightblue}}

\title{Expectation Confirmation Preference Optimization for Multi-Turn Conversational Recommendation Agent}

\author{
Xueyang Feng$^{1,2,3}$\thanks{Work done during internship at Huawei Noah's Ark Lab}, Jingsen Zhang$^{1,2,3}$, Jiakai Tang$^{1,2,3}$, Wei Li$^{1}$, Guohao Cai$^{4}$,\\
\textbf{ Xu Chen$^{1,2,3}$\thanks{Corresponding author.}, Quanyu Dai$^{4}$\footnotemark[2], Yue Zhu$^{4}$, Zhenhua Dong$^{4}$} \\
$^1$Gaoling School of Artificial Intelligence, Renmin University of China, Beijing, China \\
$^2$Beijing Key Laboratory of Research on Large Models and Intelligent Governance \\
$^3$Engineering Research Center of Next-Generation Intelligent Search and Recommendation, MOE \\
$^4$Huawei Noah's Ark Lab, China \\
\texttt{\{xueyangfeng, xu.chen\}@ruc.edu.cn} \\
\texttt{daiquanyu@huawei.com}
}

\begin{document}
\maketitle
\begin{abstract}
Recent advancements in Large Language Models (LLMs) have significantly propelled the development of Conversational Recommendation Agents (CRAs). However, these agents often generate short-sighted responses that fail to sustain user guidance and meet expectations. 
Although preference optimization has proven effective in aligning LLMs with user expectations, it remains costly and performs poorly in multi-turn dialogue.
To address this challenge, we introduce a novel multi-turn preference optimization (MTPO) paradigm \textbf{ECPO}
, which leverages Expectation Confirmation Theory to explicitly model the evolution of user satisfaction throughout multi-turn dialogues, uncovering the underlying causes of dissatisfaction.
These causes can be utilized to support targeted optimization of unsatisfactory responses, thereby achieving turn-level preference optimization.
ECPO ingeniously eliminates the significant sampling overhead of existing MTPO methods while ensuring the optimization process drives meaningful improvements.
To support ECPO, we introduce an LLM-based user simulator, \textbf{AILO}, to simulate user feedback and perform expectation confirmation during conversational recommendations. Experimental results show that ECPO significantly enhances CRA's interaction capabilities, delivering notable improvements in both efficiency and effectiveness over existing MTPO methods.
\end{abstract}


\section{Introduction}
Conversational Recommendation Systems (CRSs) leverage multi-turn natural language interactions to gradually uncover user interests and subsequently recommend items aligned with their preferences~\cite{CRS_Survey_Jannach,Gao_2021}. 
Powered by the advanced text generation and tool-calling capabilities of Large Language Models (LLMs)~\citep{Wang_2024}, LLM-based Conversational Recommendation Agents (CRAs)~\citep{gao2023chatrec, huang2023interecagent, fang2024multi} are emerging as a mainstream paradigm for delivering accurate, interpretable, and emotionally engaging personalized services.
However, the responses generated by current CRAs often appear rigid, lacking proactivity and flexibility.
This is mainly because the pretraining objectives of LLMs are predominantly focused on short-sighted next-token prediction~\citep{ouyang2022training}. As a result, their ability to sustain multi-turn interactions and provide dynamic guidance is limited, making it difficult to meet human expectations in conversation.

\begin{figure}[t]
    \centering
    \includegraphics[width=\columnwidth]{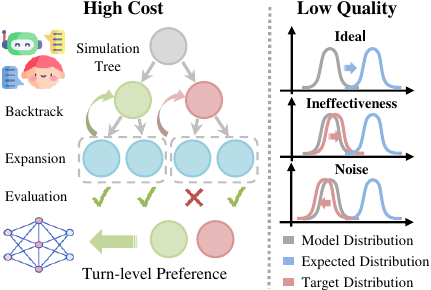}
    \caption{ Existing MTPO methods have three inherent challenges: (1) Tree simulation incurs additional sampling \textbf{Costs}. (2) In CRA tasks, LLMs struggle to generate \textbf{Effective} positive examples through self-sampling. (3) Simulated environmental \textbf{Noise} in the expansion and evaluation may be incorporated into preference relations, leading the model to update in the wrong direction.
    }
    \label{fig:intro}
\end{figure}

 To address this challenge, aligning CRAs with human expectations presents a viable solution. Preference optimization has demonstrated success in aligning LLM outputs with user preferences~\citep{schulman2017proximalpolicyoptimizationalgorithms,ouyang2022training,rafailov2024directpreferenceoptimizationlanguage}. Its core principle involves sampling multiple candidate outputs from the LLM and increasing the probability of those that align with user expectations.
 However, conversational recommendation is a multi-turn dialogue task, and applying preference optimization to this process presents great challenges. 
 The main difficulty is that user preferences change in each dialogue turn and dynamically evolve as the conversation progresses.
Most existing Multi-Turn Preference Optimization (MTPO) methods simply treat each turn equally, failing to capture turn-level preference relationships ~\citep{ulmer2024bootstrappingllmbasedtaskorienteddialogue,sun2024parrotenhancingmultiturninstruction}.
Several recent works~\citep{jin2024persuading, xie2024montecarlotreesearch} try to infer turn-level preference relationships through tree-based simulations. As illustrated in Fig.~\ref{fig:intro}, these approaches introduce three 
inherent challenges:
  (1) To obtain turn-level preference, it is necessary to sample multiple candidate responses at each turn and simulate the entire conversation to evaluate preferences for intermediate turns, resulting in \textbf{significant sampling overhead}. 
  (2) In multi-turn conversational recommendation tasks, LLMs struggle to generate \textbf{effective} positive outputs through self-sampling. (3) Evaluating preferences for intermediate turns relies on the simulated environment, whose randomness may introduce additional \textbf{noise} into preference relationships, leading to suboptimal performance of the aligned CRA.
 Overcoming these limitations is essential to aligning CRAs with human expectations. This leads to a critical question: \textit{Is there a way to construct high-quality turn-level preference relationships without additional sampling and evaluation?}

 \textit{A problem well stated is a problem half solved}. 
 The core idea of this paper is to explicitly model how user satisfaction evolves throughout multi-turn dialogues and uncover the underlying causes of dissatisfaction. 
 By identifying and addressing the root causes of low satisfaction, we can naturally construct responses that better align with user expectations. Expectation Confirmation Theory (ECT) \citep{Oliver1977, Oliver1980} tells us satisfaction is a subjective feeling that arises from the comparison between an individual's initial expectations and the perceived actual performance or outcomes. When applied to the context of conversational recommendation, this can be understood as: during a dialogue, a user has specific expectations for the system's response in each turn. Upon receiving the actual response, the user evaluates it by comparing it with their initial expectations, assigning a subjective satisfaction score based on the perceived gap.

Motivated by this, we propose \textbf{E}xpectation \textbf{C}onfirmation \textbf{P}reference \textbf{O}ptimization (\textbf{ECPO}), which comprises three key steps: 
(1) \textit{Forward Expectation Confirmation} to identify unsatisfactory responses and uncover their root causes; (2) \textit{Backward Expectation Derivation} to rewrite the unsatisfactory responses based on these causes; (3) \textit{Preference Optimization} using the original and rewritten responses.
Considering the high cost and potential bias associated with real users participating in the Expectation Confirmation (EC) process, we further introduce \textbf{AILO}, an LLM-based agent that simulates real users’ \textbf{A}ctivities, \textbf{I}nterests, \textbf{L}anguage, and \textbf{O}rientations. 
 During the dialogue, AILO acts as a user, providing diverse and realistic feedback as well as performing the EC process.
Our contributions are summarized as follows:
\begin{itemize}
\setlength{\itemsep}{0pt}
\item We introduce ECPO, a novel MTPO paradigm leveraging ECT to guide turn-level alignment in  dialogues. To the best of our knowledge, this is the first preference optimization method tailored for LLM-based CRAs.
\item To support ECPO, we introduce an LLM-based user simulator, AILO, which provides diverse and realistic feedback as well as performs the expectation confirmation process.
\item We conduct extensive experiments on three datasets, demonstrating ECPO’s exceptional performance in enhancing CRA’s interactive capabilities and highlighting its significant advantages over existing MTPO methods in both efficiency and effectiveness. 
\end{itemize}

\section{Method}
To better align multi-turn CRAs with human expectations, we propose \textbf{E}xpectation \textbf{C}onfirmation \textbf{P}reference \textbf{O}ptimization (\textbf{ECPO}).  Its core idea is to leverage ECT to explicitly model the evolution of user satisfaction throughout multi-turn dialogues and construct turn-level preference relationships by identifying and addressing the root causes of dissatisfaction. A detailed description of ECPO is provided in Section \ref{sec: ecpo}. Additionally, we introduce a novel user simulator, AILO, which generates diverse and realistic user feedback while performing expectation confirmation (see Section \ref{sec: ailo}).

\begin{figure*}[t]
    \centering
    \includegraphics[width=\textwidth]{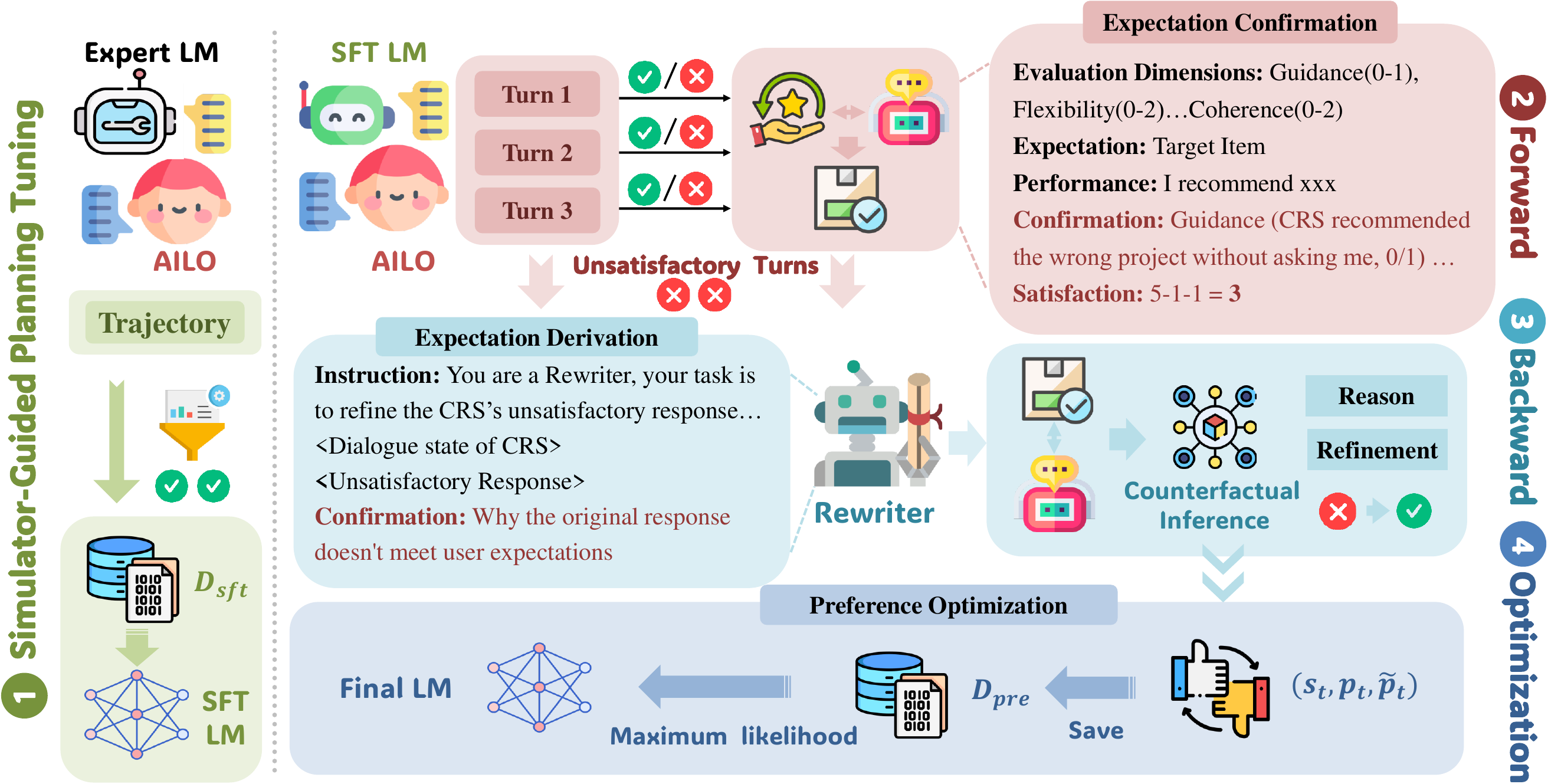}
    \caption{The ECPO process consists of four steps: (1) \textit{Simulator-Guided Planning Tuning} to distill $\pi_\text{sft}$ from the GPT-based CRA; (2) \textit{Forward Expectation Confirmation} to identify unsatisfactory responses and uncover their root causes; (3) \textit{Backward Expectation Derivation} to rewrite unsatisfactory responses based on the EC process; (4) \textit{Preference Optimization} based on the original and rewritten responses.}
    \label{fig:method}
\end{figure*}

\subsection{Preliminary}
\label{sec: pre}
We define the CRA as $\pi$\footnote{The backbone of $\pi$ is a tunable open-source LLM. In this paper, we use Llama-3.1-8B-Instruct~\citep{grattafiori2024llama3herdmodels}.}, which leverages LLMs' planning and tool-calling capabilities to conduct multi-turn dialogues with a user $U$. Through iterative interactions, the agent elicits user preferences, retrieves relevant items from the external database \(I = \{ I_1, I_2, \dots, I_n \}\), and recommends the item that best matches the user's interests. Formally, at the $t$-th turn ($1 \leq t \leq T$), $\pi$ performs internal reasoning $cr_t$ and generates a response $p_t$, denoted as $\{cr_t, p_t\} = \pi(s_t)$, where $s_t$ represents the dialogue state (e.g., dialogue history). We follow the setting proposed by iEvalLM \citep{iEvalLM}, which assumes each user has a ground-truth item $i^E$. The goal of the CRA is to proactively guide users in conversations, providing a highly flexible and coherent user experience while successfully recommending the target item $i^E$.
Formally, an interaction episode is:
\[
H^T = \bigl\{u_0, (cr_1, p_1, u_1), \dots, (cr_T, p_T, u_T)\bigr\},
\]
where $u_t$ represents the user's utterance at turn $t$.

\subsection{ECPO}
\label{sec: ecpo}
In this section, we propose \textbf{ECPO}, an MTPO paradigm based on ECT. As shown in Figure~\ref{fig:method}, we first obtain the model $\pi_{\text{sft}}$ through a \textit{Simulator-Guided Planning Tuning} phase. Subsequently, ECPO is performed in three steps: \textit{Forward Expectation Confirmation}, \textit{Backward Expectation Derivation}, and \textit{Preference Optimization}.

\paragraph{Simulation-Guided Planning Tuning.} Existing CRS datasets~\citep{kim2024pearlreviewdrivenpersonaknowledgegrounded} often lack an internal reasoning process, making them unsuitable for CRA's fine-tuning. To resolve this issue, we construct a new multi-turn conversational recommendation dataset that incorporates internal reasoning. This dataset is generated from dialogues between a GPT-4o mini-based CRA $\pi_\text{GPT}$ and a user simulator \( U \). We filter the trajectories based on whether the recommendation is successful, resulting in the dataset \( \mathcal{D}_{\text{sft}} \). Subsequently, we perform supervised fine-tuning (SFT) on the CRA $\pi$:
\begin{align}
\mathcal{L}_{\text{SFT}} = \mathbb{E}_{(s_t, cr_t, p_t) \sim \mathcal{D}_{\text{sft}}} \left[  
- \log \pi_{\theta}(cr_t, p_t | s_t)  
\right] \label{eq:sft}
\end{align}
Through this process, we obtain the CRA $\pi_{\text{sft}}$. However, SFT struggles to capture turn-level user preferences, making it insufficient to fully meet user expectations.
 To address this, we introduce ECPO, a low-cost and high-quality MTPO paradigm. For clarity, we omit the internal reasoning \( cr \) of the CRA in the subsequent formulations.

\paragraph{Forward Expectation Confirmation.}
\label{eq:ect_satisfaction}
Expectation Confirmation Theory tells us an individual's satisfaction arises from comparing actual performance against prior expectations. 
When applied to conversational recommendation, the evolution of user satisfaction can be modeled through the Expectation Confirmation (EC) process. In this paper, we adopt an extensible multi-dimensional scoring criterion with a maximum score of 5, consisting of \textit{flexibility (0-2 points), coherence (0-2 points), and user guidance ability (0-1 point)}~\citep{Gao_2021, alkan2019irfinteractiverecommendationdialogue}. 
Formally, at the $t$-th turn, ECPO integrates the user expectation item $i^E$ and the CRA's response $p_t$ at this dialogue turn into an instruction prompt $I_\text{ect}$.
 The instruction is designed to explicitly simulate the user's inner monologue during the conversation: First, a user $U$ evaluates the system's output against their expectations, assessing whether each dimension meets the corresponding requirement and assigning a sub-score to each aspect. These sub-scores are then aggregated to compute the overall satisfaction score $r_t$ for $p_t$. We formulate the EC process as follows: 
\[\{\text{CONF}_t, r_t\} = U(I_\text{ect}(i^E, h_t, p_t)),\]
where $h_t$ is the dialogue history, $\text{CONF}_t$ is a natural language explanation explicitly detailing why the user feels satisfied or dissatisfied at this turn.
We then trace back the internal state \( s_t \) at the time of the CRS output \( p_t \), together with the corresponding EC process $\text{CONF}_t$, and store it as a tuple \( (s_t, p_t, \text{CONF}_t, r_t) \) for the subsequent phase.


\begin{figure*}[t]
    \centering
    \includegraphics[width=\textwidth, height=4.8cm]{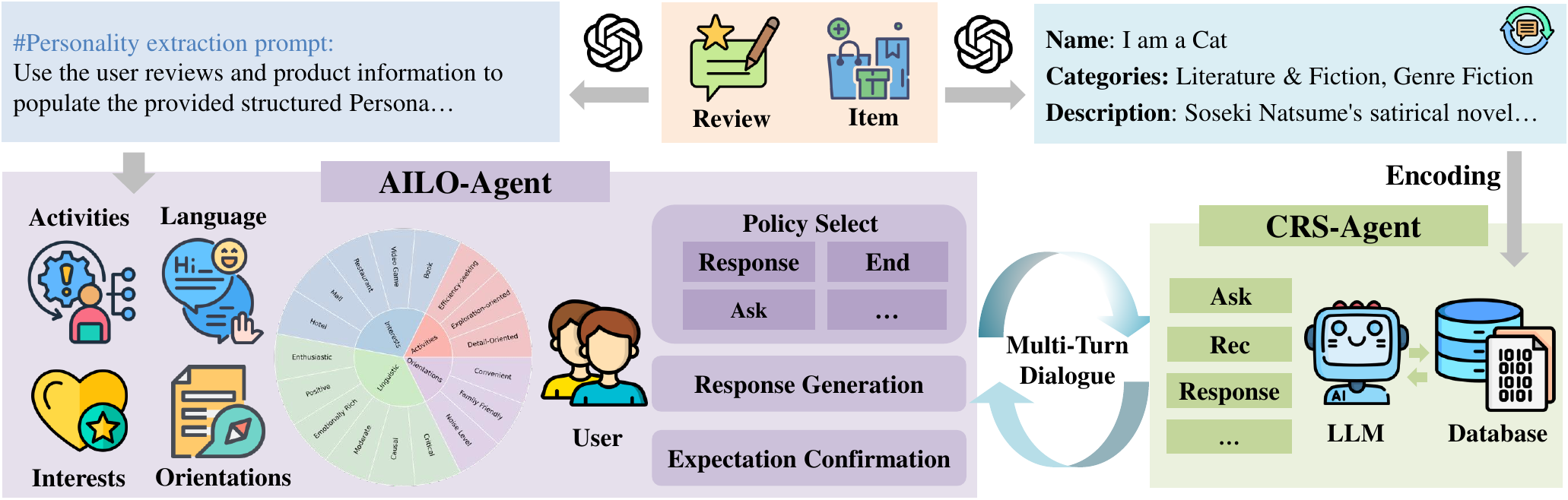}
    \caption{The illustration of the AILO, showing its \textit{persona modeling}, \textit{policy-based user simulation}. Figure also depicts the task of the CRA: interacting with the database, engaging in dialogue, and recommending items to AILO.}
    \label{fig:ailo}
\end{figure*}

\paragraph{Backward Expectation Derivation.}
Once each dialogue turn is assigned a satisfaction score via the EC process, we can identify responses that fail to meet user expectations.
Next, we backtrack to the CRA state $s_t$ and leverage $\text{CONF}_t$ for counterfactual inference on how the CRA should have generated a response to better align with user expectations. Formally, at the $t$-th turn, ECPO integrates the EC process $\text{CONF}_t$ and the unsatisfactory response $p_t$ into an instruction prompt $I_\text{bed}
$, which serves as the input for the Rewriter—an additional LLM introduced to refine unsatisfactory responses during backtracking.
The Rewriter employs a slow thinking process, first generating a chain of thought~\citep{wei2023chainofthoughtpromptingelicitsreasoning} and then producing a refined response $\tilde{p}_t$:

\begin{equation}
    \label{eq:refinement}
    \begin{aligned}
        \tilde{p}_t &= \text{Rewriter}(I_\text{bed}(s_t, p_t, \text{CONF}_t)), \\
        &\quad \text{where } r_t \leq \lambda
    \end{aligned}
\end{equation}
Here, $\lambda$ is a hyperparameter that defines the satisfaction threshold. If the user's satisfaction score $r_t$ falls below $\lambda$, the response will undergo backtracking and rewriting. 
Meanwhile, to ensure that rewritten responses do not deviate too far from the $\pi_\text{sft}$, we require the Rewriter to make only limited modifications to the unsatisfactory response, rather than performing a complete rewrite. 

After the backward process, we can collect these ``original--rewritten'' pairs from the training set to form our preference dataset, denoted as \( \mathcal{D}_{\text{pre}} = \{(s_t, p_t, \tilde{p}_t) \mid r_t < \lambda \} \). This dataset consists of turn-level preference pairs, where the rewritten responses $\tilde{p}_t$ are statistically more likely to exhibit significant improvements over the original ones. This hypothesis has been empirically validated through our evaluation (cf. Appendix \ref{sec: emofrewrite}).

\paragraph{Preference Optimization.}
After obtaining the turn-level preference dataset $\mathcal{D}_{\text{pre}}
$, we can optimize $\pi_{sft}$ through existing preference optimization methods.
A typical implementation is Direct Preference Optimization (DPO)~\citep{rafailov2024directpreferenceoptimizationlanguage}:
\begin{multline}
\mathcal{L}_{\text{DPO}}(\pi_\theta, \pi_{\text{sft}}) = \mathbb{E}_{s, \tilde{p}_t, p_t \sim \mathcal{D_\text{pre}}}   \bigg[ 
-\log \sigma  \\ \bigg(
\beta \log \frac{\pi_\theta(\tilde{p}_t \mid s_t)}{\pi_{\text{sft}}(\tilde{p}_t \mid s_t)}
- \beta \log \frac{\pi_\theta(p_t \mid s_t)}{\pi_{\text{sft}}(p_t \mid s_t)}
\bigg) \bigg]
\end{multline}
ECPO is both orthogonal and complementary to existing preference optimization methods. This enables seamless integration with various methods (e.g., KTO~\citep{ethayarajh2024ktomodelalignmentprospect}, SimPO~\citep{meng2024simposimplepreferenceoptimization}) based on specific task requirements and optimization goals. 

\paragraph{Discussion}
Existing MTPO methods typically require completing the entire conversation before estimating the reward for each intermediate turn, and all positive samples must be generated through self-sampling. In contrast, ECPO implicitly assigns rewards at each turn through the EC process and provides the underlying reasons for these rewards in natural language. 
These reasons promote the proactive generation of positive samples for preference optimization instead of self-sampling.
This paradigm not only eliminates additional sampling and evaluation costs but also ensures that preference relationships drive meaningful optimization. In the next section, we introduce AILO, a novel user simulator designed to support the EC process.

\subsection{AILO}
\label{sec: ailo}
This paper aims to leverage the EC process to explicitly model how user satisfaction evolves throughout conversational recommendation, thereby guiding CRA to align with user expectations. However, considering the unacceptably high costs and potential biases involved in human participation, we propose a new user simulator, \textbf{AILO}, an LLM-based agent that provides realistic and diverse user feedback. As shown in Figure~\ref{fig:ailo}, AILO consists of two components: \textit{user persona modeling} and \textit{policy-based user simulation}.

\begin{figure}[t]
    \centering
    \includegraphics[width=\columnwidth]{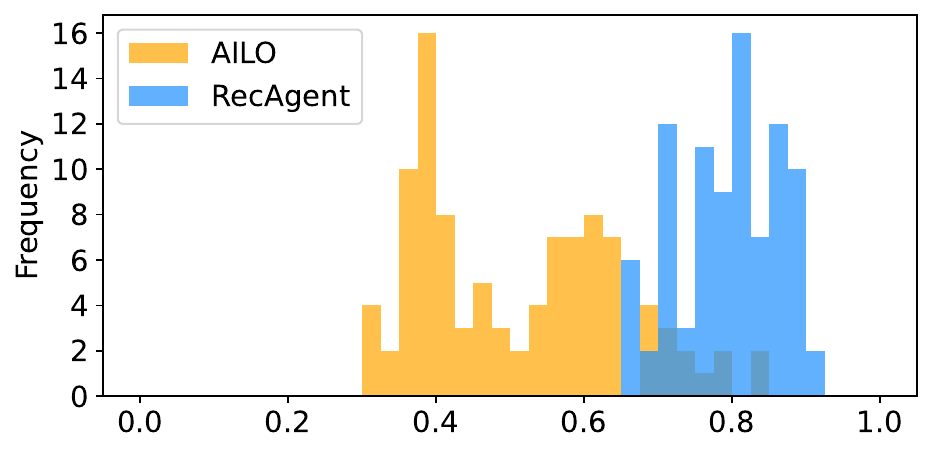}
    \caption{ROUGE-L with the Most Similar Persona.}
    \label{fig:rouge}
\end{figure}

\paragraph{User Persona Modeling.}
Existing user simulators typically generate user personas through simple random sampling~\citep{wang2024userbehaviorsimulationlarge}, but this approach often results in unrealistic and less diverse personas.
To address this, we propose \textbf{AILO}, a comprehensive user simulator for conversational recommendation. Inspired by the AIO theory~\citep{wells1971opinions} from consumer psychology, AILO defines user attributes across four dimensions: Activities, Interests, Language, and Orientations, thereby capturing the diverse characteristics that users may exhibit during conversational recommendations. For example, some users prioritize efficiency in recommendations, while others prefer engaging in in-depth discussions on specific topics. 
We employ GPT-4o~\citep{openai2024gpt4technicalreport} to infer user personas from real recommendation review datasets. This not only ensures the authenticity of personas but also enhances their diversity. To assess the diversity of AILO's personas, following~\citet{jin2024persuading}, we randomly sample 100 personas created by our method and those generated using the sampling method in RecAgent~\citep{wang2024userbehaviorsimulationlarge}, then compute the maximum ROUGE-L between each persona and the others. As shown in Figure~\ref{fig:rouge}, the ROUGE-L's distribution of AILO is significantly lower than RecAgent, indicating that AILO produces more diverse user personas. 

\paragraph{Policy-Based User Simulation.}
Directly simulating user responses with LLMs may lead to role reversals and uncontrollable behavior~\citep{zhu2024llmbasedcontrollablescalablehumaninvolved}. 
Therefore, we redefine the process of user response generation as a planning task executed in three steps: (1) \textit{Response Policy Generation}: Based on the user's persona and the CRA's response $p_t$, the simulator $U$ generates a response policy $ur_t$, such as ``Asking for Recommendations''. (2) \textit{Response Content Generation}: Based on the response policy $ur_t$, the simulator generates the response $u_t$. (3) \textit{Expectation Confirmation Process}: $U$ generates the EC process $\text{CONF}_t$, computes the satisfaction score \( r_t \), and outputs them in a structured format.
Formally, the simulator produces:
\begin{equation}
    \{ur_t, u_t, \text{CONF}_t, r_t\} = U(i^E, h_t, p_t)
\end{equation}
Here, \( i^E \) is the target item, and \( h_t \) represents the dialogue history. 
To verify the authenticity of AILO's simulated dialogue, we recruit annotators to compare 50 sets of dialogue trajectories generated by AILO and iEvalLM~\citep{iEvalLM}, assessing which one appears more human-like. The experimental results show that AILO outperforms iEvalLM in all cases, achieving a 100\% win rate.

\section{Experiments}

\begin{table*}[t]
\centering
\setlength{\tabcolsep}{4.55pt}  
\begin{tabular}{llcccccccccc}
\toprule
\multirow{2}{*}{\textbf{Backbone}} & \multirow{2}{*}{\textbf{Method}} & \multirow{2}{*}{\textbf{\#Calls}}
& \multicolumn{3}{c}{\textbf{Book}} 
& \multicolumn{3}{c}{\textbf{Game}}
& \multicolumn{3}{c}{\textbf{Yelp}} \\
\cmidrule(lr){4-6} \cmidrule(lr){7-9} \cmidrule(lr){10-12}
& & & \textbf{SR} & \textbf{R} & \textbf{WR} 
& \textbf{SR} & \textbf{R} & \textbf{WR}
& \textbf{SR} & \textbf{R} & \textbf{WR} \\
\midrule
\multirow{4}{*}{\shortstack{GPT-4o \\ mini}} 
& ChatRec & $\mathcal{O}(N)$
& 0.46 & 0.47 & 0.13 
& 0.37 & 0.45 & 0.09
& 0.24 & 0.30 & 0.12 \\
& ReAct & $\mathcal{O}(M+2N)$
& 0.52 & 0.56 & 0.33
& 0.39 & 0.65 & 0.34
& 0.57 & 0.62 & 0.42 \\
& MACRS & $\mathcal{O}(M+4N)$
& 0.63 & 0.71 & 0.01
& 0.36 & 0.65 & 0.15
& 0.40 & 0.41 & 0.02 \\
& ActCRS & $\mathcal{O}(M+N)$
& 0.53 & 0.56 & 0.50 
& 0.43 & 0.68 & 0.50
& 0.37 & 0.43 & 0.50 \\
\midrule
\multirow{6}{*}{\shortstack{Llama-3.1 \\ 8B-Instruct}}
& ChatRec & $\mathcal{O}(N)$
& 0.42 & 0.47 & 0.03 
& 0.36 & 0.39 & 0.01 
& 0.30 & 0.32 & 0.05 \\
& ReAct & $\mathcal{O}(M+N)$
& 0.36 & 0.54 & 0.19
& 0.04 & 0.43 & 0.08
& 0.31 & 0.40 & 0.16 \\
& MACRS & $\mathcal{O}(M+4N)$
& 0.36 & 0.39 & 0.01
& 0.24 & 0.34 & 0.00
& 0.22 & 0.24 & 0.01 \\
& ActCRS & $\mathcal{O}(M+N)$
& 0.34 & 0.55 & 0.28  
& 0.07 & 0.50 & \underline{0.46}
& 0.22 & 0.35 & 0.38 \\
& \blue{\textit{+SGPT(Ours)}} & \blue{\textbf{$\mathcal{O}(M+N)$}}
& \blue{\textbf{0.54}} & \blue{\underline{0.57}} & \blue{\underline{0.48}}
& \blue{\textbf{0.41}} & \blue{\textbf{0.61}} & \blue{0.42}
& \blue{\underline{0.44}} & \blue{\textbf{0.48}} & \blue{\underline{0.47}} \\
& \blue{\textit{+ECPO(Ours)}} & \blue{\textbf{$\mathcal{O}(M+N)$}}
& \blue{\textbf{0.56}} & \blue{\textbf{0.58}} & \blue{\textbf{0.57}}
& \blue{\textbf{0.41}} & \blue{\textbf{0.61}} & \blue{\textbf{0.56}}
& \blue{\textbf{0.45}} & \blue{\underline{0.47}} & \blue{\textbf{0.63}} \\
\bottomrule
\end{tabular}
\caption{Comparison with existing prompt-based CRAs. The "\#Calls" column represents the number of LLM calls required to complete an entire dialogue. \(N\) denotes the number of dialogue turns, and \(M\) represents the number of times the LLM generates retrieval queries (\(M \leq N\)). \textbf{SR} (Success Rate) and \textbf{R} (Recall Rate) are recommendation metrics, while \textbf{WR} reflects the interactive capabilities.
}
\label{tab:exp1}
\end{table*}

To thoroughly evaluate the effectiveness of ECPO in enhancing multi-turn CRAs, we conduct extensive experiments, which are outlined as follows:
\begin{itemize}
    \item First, to validate the importance of ECPO alignment for CRAs, we compare existing prompt-based CRAs with those that have undergone ECPO alignment.
    \item Second, we comprehensively compare ECPO with existing MTPO methods to verify its efficiency and effectiveness.
    \item Finally, we thoroughly analyze the effectiveness of different components of ECPO and conduct evaluations of its performance under various experimental settings.
\end{itemize}

\subsection{Experimental Setup}
In this section, we briefly introduce the experimental settings. A more detailed elaboration and design motivations are presented in Appendix~\ref{app: exp_details}.

\paragraph{Environments.} Traditional CRS evaluation methods struggle to assess dynamic CRA tasks~\citep{Afzali_2023}. As discussed in Section~\ref{sec: ailo}, we follow and extend iEvalLM~\citep{iEvalLM} by introducing AILO for our evaluations. Our experiments utilize the Amazon-Game, Amazon-Book\footnote{\url{https://github.com/hyp1231/AmazonReviews2023}}, and Yelp\footnote{\url{https://www.yelp.com/dataset}} datasets to construct user personas and generate approximately 3,000 tasks for each dataset. During the training phase, we use 1,000 tasks to construct \(\mathcal{D}_{\text{sft}}\) and 500 tasks to construct \(\mathcal{D}_{\text{pre}}\). Following ReAct~\citep{yao2023react} and MACRS\citep{fang2024multi}, we sample 100 tasks from each dataset for testing. 

\paragraph{Baselines.}Given the significant gap between traditional CRS and emerging LLM-based CRAs, we focus on comparing our approach with existing prompt-based CRAs (ChatRec~\citep{gao2023chatrec}, ReAct~\citep{yao2023react}, MACRS~\citep{fang2024multi}, ActCRS) and MTPO methods (trajectory-level: SFT, KTO~\citep{ethayarajh2024ktomodelalignmentprospect}; turn-level: SDPO~\citep{jin2024persuading}, SKTO). Notably, ActCRS is a straightforward CRA developed by us, that simultaneously generates a response strategy and the corresponding response. Due to its simplicity and effectiveness, we fine-tune ActCRS in our main experiments.
Our backbone model is Llama-3.1-8B-Instruct~\citep{grattafiori2024llama3herdmodels}, and we additionally provide results based on GPT-4o mini~\citep{openai2024gpt4technicalreport} as a reference.

\begin{figure*}[ht!]
    \centering
    \includegraphics[width=\textwidth]{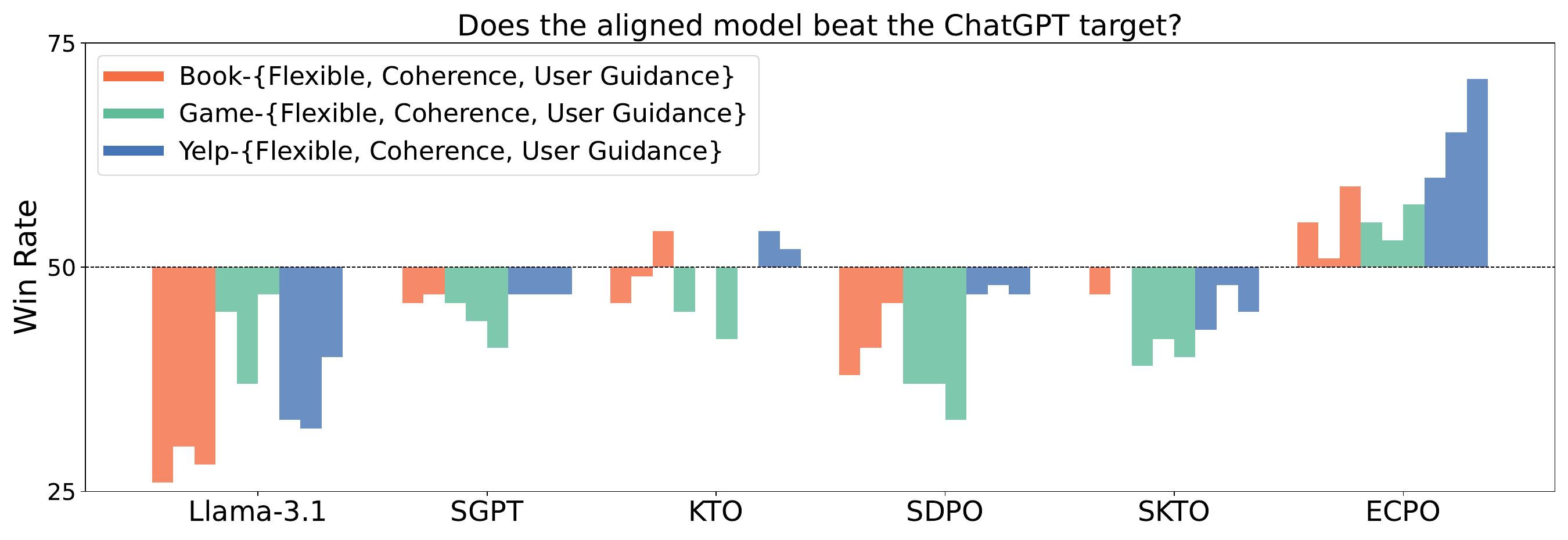}
    \caption{Comparison of aligned CRAs fine-tuned with different methods in terms of interactivity (flexibility, coherence, and user guidance) against the expert CRA.}
    \label{fig:exp2}
    \vspace{-1pt}
\end{figure*}

\begin{figure}[h]
    \centering
    \includegraphics[width=\columnwidth]{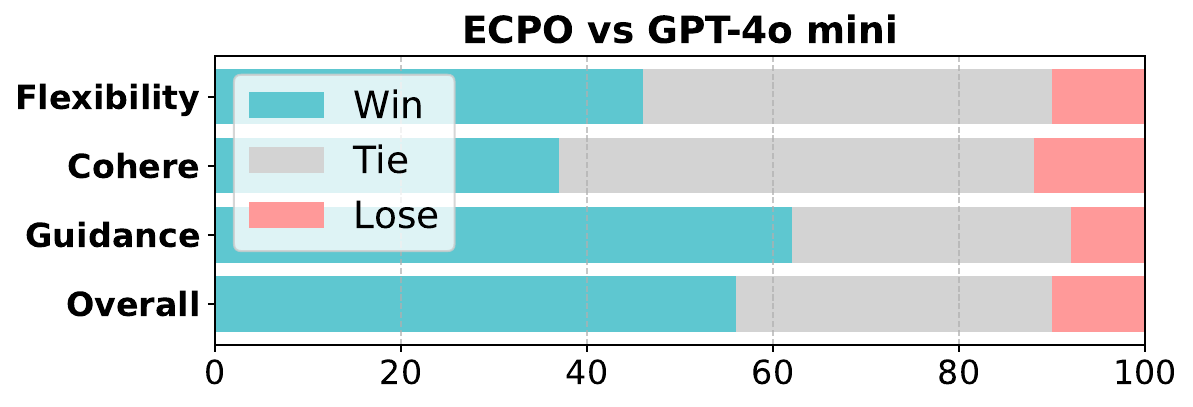}
    \caption{Human evaluation results.}
    \label{fig:human}
    \vspace{-5pt}
\end{figure}

\paragraph{Metrics.} We evaluate CRAs across two dimensions:  
(1) \textbf{Recommendation Metrics}: Success Rate (\textbf{SR}) and Recall Rate (\textbf{R}).  
(2) \textbf{Dialogue Metric}: Win Rate (\textbf{WR}, ~\citep{alpaca_eval}), which measures interactivity compared to the expert CRA  (GPT-based ActCRS in main experiments).

\subsection{Comparison with Existing Prompt-Based CRA Frameworks}
\paragraph{Analysis of Existing Prompt-Based CRAs.}Table \ref{tab:exp1} summarizes the main experimental results on three recommendation datasets. First, we analyze the existing CRAs' results. We find that:
(1) Stronger backbone models (GPT-4o mini) perform better as CRA framework complexity increases. In contrast, weaker models (Llama-3.1) struggle to benefit from more complex CRA frameworks.
(2) ChatRec and MACRS can generate high-quality recommendations. However, ChatRec lacks interactivity, while MACRS's responses tend to be overly verbose, making conversations feel unnatural. In terms of \textbf{WR} (interactivity performance), their win rates are significantly lower than expert CRA, typically below 0.15.
(3) No single prompt-based CRA demonstrates a clear advantage across all datasets and metrics. Moreover, as the number of calls increases, the performance gains gradually diminish. This observation highlights the growing importance of an alignment method for CRAs.

\paragraph{Effect of Alignment.}We fine-tune the Llama-based ActCRS using SGPT + ECPO, and present the performance results in the table~\ref{tab:exp1}. After SGPT training, the recommendation metrics (\textbf{SR} and \textbf{R}) reach GPT-level performance, but interactivity remains inferior to the expert CRA. After ECPO training, the win rate significantly exceeded that of the GPT model (\textbf{WR} ranging from 0.56 to 0.63), highlighting the crucial role of the ECPO in enhancing the multi-turn conversation user experience.

\begin{table*}[t]
\centering
\setlength{\tabcolsep}{8pt}
\renewcommand{\arraystretch}{1.15}
\resizebox{\textwidth}{!}{
\fontsize{9}{10}\selectfont
\begin{tabular}{l *{9}{c}} 
\toprule
\noalign{\vskip -1pt}
\multirow{2}{*}{\textbf{Method}}
& \multicolumn{3}{c}{\textbf{Book}}
& \multicolumn{3}{c}{\textbf{Game}}
& \multicolumn{3}{c}{\textbf{Yelp}} \\
\noalign{\vskip -2pt}
\cmidrule(lr){2-4} \cmidrule(lr){5-7} \cmidrule(lr){8-10} 
\noalign{\vskip -2pt}
& \textbf{SR} & \textbf{R} & \textbf{WR}
& \textbf{SR} & \textbf{R} & \textbf{WR}
& \textbf{SR} & \textbf{R} & \textbf{WR} \\
\noalign{\vskip -2pt} 
\midrule
\noalign{\vskip -1pt}
Llama-3.1
& 0.34 & 0.55 & 0.28 
& 0.07 & 0.50 & 0.46 
& 0.22 & 0.35 & 0.38 \\
\textit{+SGPT}
& \underline{0.54} & \underline{0.57} & 0.48 & \textbf{0.41} & \textbf{0.61} & 0.42
& \underline{0.44} & \textbf{0.48} & 0.47 \\
\blue{\textit{+ECPO-w/o EC}}
& \blue{0.50} & \blue{0.53} & \blue{\underline{0.54}} & \blue{\underline{0.37}} & \blue{\underline{0.55}} & \blue{\underline{0.54}}
& \blue{0.42} & \blue{0.46} & \blue{\underline{0.48}} \\ 
\blue{\textit{+ECPO}}
& \blue{\textbf{0.56}} & \blue{\textbf{0.58}} & \blue{\textbf{0.57}}
& \blue{\textbf{0.41}} & \blue{\textbf{0.61}} & \blue{\textbf{0.55}}
& \blue{\textbf{0.45}} & \blue{\underline{0.47}} & \blue{\textbf{0.63}} \\
\noalign{\vskip -2pt}
\bottomrule
\end{tabular}
}
\caption{Effectiveness of the EC process.}
\label{tab:eec}
\end{table*}

\begin{table}[ht]
\centering
\begin{tabular}{lccc}
\toprule
\textbf{Metric} & \textbf{Book} & \textbf{Game} & \textbf{Yelp} \\
\midrule
Information Fidelity & 0.64 & 0.64 & 0.58 \\
Coherence            & 0.90 & 0.82 & 0.73 \\
\bottomrule
\end{tabular}
\caption{Win rate of Rewritten vs. Original responses across Fidelity \& Coherence.}
\label{tab:bias}
\end{table}

\subsection{Comparison with Existing MTPO Methods}
\label{sec:exp2}
In Figure~\ref{fig:exp2}, we compare ECPO with two categories of existing multi-turn alignment methods: trajectory-level methods (SFT, KTO) and turn-level preference optimization methods based on tree simulation (SDPO, SKTO). Specifically, we construct the preference dataset $\mathcal{D}_{\text{pre}}$ using each method in 500 simulation tasks. In these tasks, trajectory-level methods require sampling 1,000 trajectories, tree simulation methods require sampling 2,500 trajectories, whereas ECPO eliminates the need for additional sampling and efficiently utilizes only 500 trajectories.
Experimental results show that the improvement of trajectory-level methods is limited, as they fail to effectively capture preference relationships at the turn level. Meanwhile, tree simulation methods, despite capturing these preferences, actually led to negative gains, likely due to noise interference.
This finding highlights the challenges of CRA alignment. In contrast, ECPO, guided by the EC process, achieves the best performance while requiring the lowest cost, significantly outperforming all existing methods.

Additionally, we recruit human annotators to compare the win rates between the ECPO-aligned CRA and the expert CRA. The experimental results, as shown in Figure~\ref{fig:human}, indicate that ECPO demonstrates a significant advantage across all metrics, especially in flexibility and user guidance. To further understand how ECPO outperforms existing methods, we provide case studies in appendix~\ref{sec: case}. 


\subsection{Effectiveness of the EC Process}
Although we have demonstrated the effectiveness of ECPO in the main experiments, a natural question arises: \textit{How does the turn-level EC process influence the performance of ECPO?} To investigate this further, we manually design rewriting instructions based on the test results of $\pi_\text{sft}$, identifying its issues and guiding the Rewriter to revise the responses generated by $\pi_\text{sft}$, to construct $\mathcal{D}_\text{pre}$. This approach, referred to as ECPO w/o EC, aims to replace each turn of the EC process with a unified analysis conducted by human to guide rewriting.

In Table~\ref{tab:eec}, we find that ECPO w/o EC enhances interactivity to some extent but slightly reduces recommendation performance, with overall performance remaining significantly inferior to ECPO. This result underscores the importance of the turn-level EC process in the rewriting process.


\subsection{Hyperparameter Analysis}
\begin{figure}[t]
    \centering
    \includegraphics[width=\columnwidth]{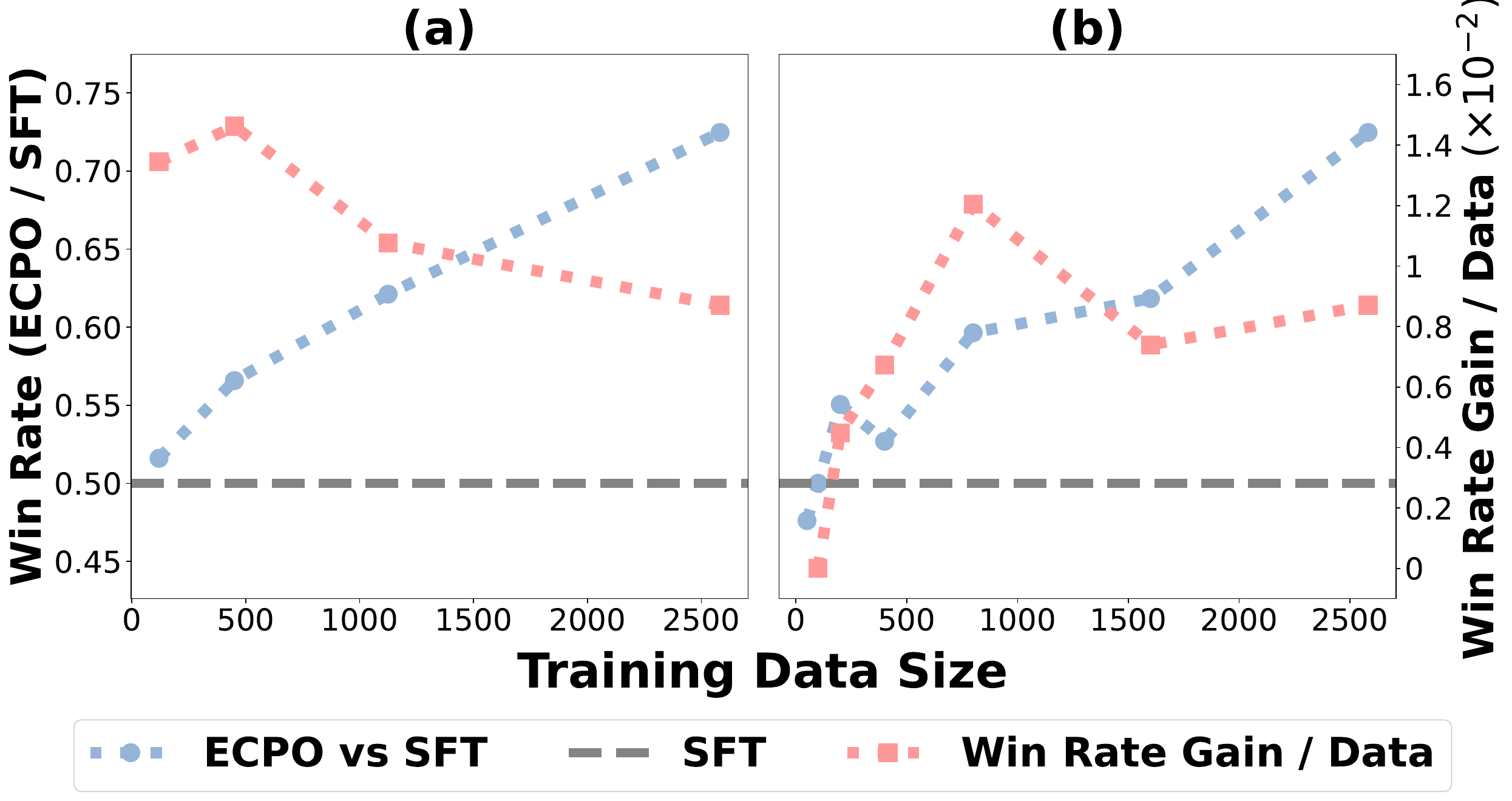}
    \caption{Hyperparameter analysis of $\lambda$.}
    \label{fig:hyper}
\end{figure}

In this section, we investigate the impact of the rewriting threshold $\lambda$, defined as the satisfaction score threshold below which responses are selected for rewriting and training. A higher $\lambda$ leads to more response samples being backtracked and rewritten, resulting in a larger training dataset. Figure~\ref{fig:hyper}(a) presents the training results for $\lambda$ values \{1, 2, 3, 4\}, while Figure~\ref{fig:hyper}(b) shows results from uniformly sampled subsets of the $\lambda=4$ setting with varying sample sizes \{50, 100, 200, 400, 800, 1600, All\}.

The blue line represents the overall performance gain, while the pink line represents the performance improvement per individual sample. We observe that, in Figure~4(a), lower $\lambda$ values lead to a more significant gain for individual samples. In contrast, in Figure~4(b), the performance improvement appears more irregular. This phenomenon is particularly interesting and aligns with intuition: when a sample has a lower satisfaction score, it often indicates critical issues, and addressing these issues results in a more noticeable performance gain.

\subsection{Bias Analysis of Rewritten Outputs}
In this section, we conduct an empirical analysis of potential biases introduced by the rewriter, focusing on its impact on information fidelity and conversational coherence in multi-turn dialogues. We evaluate responses across multiple domains using GPT-4o to compare the quality of rewritten versus original outputs along these two dimensions.

As shown in Table~\ref{tab:bias}, the results show that rewritten responses are generally preferred over the original ones across all domains and evaluation aspects, with particularly notable improvements in coherence. This indicates that the rewriter not only enhances language quality but also preserves the core semantic content, without introducing significant negative bias.

\section{Conclusion and Future Works}
In this work, we propose ECPO, a novel MTPO paradigm designed to enhance the interaction capabilities of LLM-based CRAs, thereby improving user satisfaction. Our core design principle is to explicitly model the evolution of user satisfaction in multi-turn dialogues and achieve turn-level preference alignment by identifying and addressing the root causes of dissatisfaction. To support ECPO, we introduce a new LLM-based user simulator, that enable more diverse and realistic simulations as well as expectation confirmation. Extensive experiments on three recommendation datasets fully demonstrate the superiority of our proposals.

Although ECPO is designed for CRAs, we believe it can be extend to broader dialogue assistants by modeling the EC process across different domains. Another promising direction is enabling LLMs to generate simulated user expectations. This capability can establish an internal feedback mechanism, integrating the EC process into the reasoning phase to help O1/R1-style~\citep{deepseekai2025deepseekr1incentivizingreasoningcapability} dialogue assistants in refining responses and further enhancing user satisfaction.

\section*{Limitations}
ECPO is a novel MTPO paradigm that performs turn-level preference optimization by simulating the dynamic evolution of user satisfaction across multi-turn conversations.
 However, the optimization process of ECPO largely relies on user simulation. To address this, we introduce AILO,a user simulator constructed from real user reviews to approximate a realistic and diverse user distribution as closely as possible. 
 Although experimental results demonstrate that AILO outperforms existing user simulators in terms of authenticity and diversity, an inevitable gap may still exist between simulated and real users. This gap may lead to distribution shift issues in real-world scenarios.
 
Nevertheless, (1) although ECPO employs AILO as a user simulator to guide CRA alignment, the aligned CRA, when evaluated by real users, still exhibits significant advantages; (2) in dynamic multi-turn recommendation scenarios, user simulator has become a key concern for both academia and industry. Therefore, we believe that ECPO remains a significant contribution to multi-turn CRA alignment, and AILO also represents a valuable contribution to CRA evaluation.

\section*{Ethical Considerations}
LLM-based multi-turn CRAs hold great potential in providing accurate, interpretable, and emotionally aware personalized recommendations. However, their development also raises ethical concerns, including potential biases, unfairness, privacy risks, and the reinforcement of filter bubbles. To mitigate these risks, we design AILO to represent a diverse range of users, aiming to reduce biases in ECPO alignment. Furthermore, we emphasize that all applications must operate under human supervision and oversight to ensure transparency and accountability. By maintaining a careful balance between technological advancement and social responsibility, we strive to foster the development of responsible AI systems.

\section*{Acknowledgments}
This work is supported in part by National Natural Science Foundation of China (No. 62422215 and No. 62472427), Major Innovation \& Planning Interdisciplinary Platform for the ``DoubleFirst Class'' Initiative, Renmin University of China, Public Computing Cloud, Renmin University of China, fund for building world-class universities (disciplines) of Renmin University of China, and Huawei Innovation Research Programs. We gratefully acknowledge the support from Mindspore \footnote{\url{https://www.mindspore.cn}}, CANN (Compute Architecture for Neural Networks) and Ascend AI Processor used for this research.

\bibliography{custom}

\begin{thebibliography}{38}
\providecommand{\natexlab}[1]{#1}

\bibitem[{Afzali et~al.(2023)Afzali, Drzewiecki, Balog, and Zhang}]{Afzali_2023}
Jafar Afzali, Aleksander~Mark Drzewiecki, Krisztian Balog, and Shuo Zhang. 2023.
\newblock \href {https://doi.org/10.1145/3539597.3573029} {Usersimcrs: A user simulation toolkit for evaluating conversational recommender systems}.
\newblock In \emph{Proceedings of the Sixteenth ACM International Conference on Web Search and Data Mining}, WSDM ’23. ACM.

\bibitem[{Alkan et~al.(2019)Alkan, Mattetti, Daly, Botea, and Vejsbjerg}]{alkan2019irfinteractiverecommendationdialogue}
Oznur Alkan, Massimiliano Mattetti, Elizabeth~M. Daly, Adi Botea, and Inge Vejsbjerg. 2019.
\newblock \href {https://arxiv.org/abs/1910.03040} {Irf: Interactive recommendation through dialogue}.
\newblock \emph{Preprint}, arXiv:1910.03040.

\bibitem[{DeepSeek-AI et~al.(2025)DeepSeek-AI, Guo, Yang, Zhang, and et~al.}]{deepseekai2025deepseekr1incentivizingreasoningcapability}
DeepSeek-AI, Daya Guo, Dejian Yang, Haowei Zhang, and et~al. 2025.
\newblock \href {https://arxiv.org/abs/2501.12948} {Deepseek-r1: Incentivizing reasoning capability in llms via reinforcement learning}.
\newblock \emph{Preprint}, arXiv:2501.12948.

\bibitem[{Ethayarajh et~al.(2024)Ethayarajh, Xu, Muennighoff, Jurafsky, and Kiela}]{ethayarajh2024ktomodelalignmentprospect}
Kawin Ethayarajh, Winnie Xu, Niklas Muennighoff, Dan Jurafsky, and Douwe Kiela. 2024.
\newblock \href {https://arxiv.org/abs/2402.01306} {Kto: Model alignment as prospect theoretic optimization}.
\newblock \emph{Preprint}, arXiv:2402.01306.

\bibitem[{Fang et~al.(2024)Fang, Gao, Ren, Chen, Verberne, and Ren}]{fang2024multi}
Jiabao Fang, Shen Gao, Pengjie Ren, Xiuying Chen, Suzan Verberne, and Zhaochun Ren. 2024.
\newblock A multi-agent conversational recommender system.
\newblock \emph{arXiv preprint arXiv:2402.01135}.

\bibitem[{Gao et~al.(2021)Gao, Lei, He, de~Rijke, and Chua}]{Gao_2021}
Chongming Gao, Wenqiang Lei, Xiangnan He, Maarten de~Rijke, and Tat-Seng Chua. 2021.
\newblock \href {https://doi.org/10.1016/j.aiopen.2021.06.002} {Advances and challenges in conversational recommender systems: A survey}.
\newblock \emph{AI Open}, 2:100–126.

\bibitem[{Gao et~al.(2023)Gao, Sheng, Xiang, Xiong, Wang, and Zhang}]{gao2023chatrec}
Yunfan Gao, Tao Sheng, Youlin Xiang, Yun Xiong, Haofen Wang, and Jiawei Zhang. 2023.
\newblock Chat-rec: Towards interactive and explainable llms-augmented recommender system.
\newblock \emph{arXiv preprint arXiv:2303.14524}.

\bibitem[{Grattafiori et~al.(2024)Grattafiori, Dubey, and et~al.}]{grattafiori2024llama3herdmodels}
Aaron Grattafiori, Abhimanyu Dubey, and et~al. 2024.
\newblock \href {https://arxiv.org/abs/2407.21783} {The llama 3 herd of models}.
\newblock \emph{Preprint}, arXiv:2407.21783.

\bibitem[{He et~al.(2023)He, Xie, Jha, Steck, Liang, Feng, Majumder, Kallus, and McAuley}]{he2023zscrs}
Zhankui He, Zhouhang Xie, Rahul Jha, Harald Steck, Dawen Liang, Yesu Feng, Bodhisattwa~Prasad Majumder, Nathan Kallus, and Julian McAuley. 2023.
\newblock Large language models as zero-shot conversational recommenders.
\newblock In \emph{Proceedings of the 32nd ACM international conference on information and knowledge management}, pages 720--730.

\bibitem[{Hu et~al.(2021)Hu, Shen, Wallis, Allen-Zhu, Li, Wang, Wang, and Chen}]{hu2021lora}
Edward~J. Hu, Yelong Shen, Phillip Wallis, Zeyuan Allen-Zhu, Yuanzhi Li, Shean Wang, Lu~Wang, and Weizhu Chen. 2021.
\newblock \href {https://arxiv.org/abs/2106.09685} {Lora: Low-rank adaptation of large language models}.
\newblock \emph{Preprint}, arXiv:2106.09685.

\bibitem[{Huang et~al.(2023)Huang, Lian, Lei, Yao, Lian, and Xie}]{huang2023interecagent}
Xu~Huang, Jianxun Lian, Yuxuan Lei, Jing Yao, Defu Lian, and Xing Xie. 2023.
\newblock Recommender ai agent: Integrating large language models for interactive recommendations.
\newblock \emph{arXiv preprint arXiv:2308.16505}.

\bibitem[{Jannach et~al.(2021)Jannach, Manzoor, Cai, and Chen}]{CRS_Survey_Jannach}
Dietmar Jannach, Ahtsham Manzoor, Wanling Cai, and Li~Chen. 2021.
\newblock A survey on conversational recommender systems.
\newblock \emph{ACM Comput. Surv.}, 54(5).

\bibitem[{Jin et~al.(2024)Jin, Ren, Kong, Wang, Song, and Chen}]{jin2024persuading}
Chuhao Jin, Kening Ren, Lingzhen Kong, Xiting Wang, Ruihua Song, and Huan Chen. 2024.
\newblock Persuading across diverse domains: a dataset and persuasion large language model.
\newblock In \emph{Proceedings of the 62nd Annual Meeting of the Association for Computational Linguistics (Volume 1: Long Papers)}, pages 1678--1706.

\bibitem[{Kim et~al.(2024)Kim, Kim, Kim, woo Kwak, Chun, Kim, Kang, Yu, Yeo, and Lee}]{kim2024pearlreviewdrivenpersonaknowledgegrounded}
Minjin Kim, Minju Kim, Hana Kim, Beong woo Kwak, Soyeon Chun, Hyunseo Kim, SeongKu Kang, Youngjae Yu, Jinyoung Yeo, and Dongha Lee. 2024.
\newblock \href {https://arxiv.org/abs/2403.04460} {Pearl: A review-driven persona-knowledge grounded conversational recommendation dataset}.
\newblock \emph{Preprint}, arXiv:2403.04460.

\bibitem[{Kong et~al.(2024)Kong, Fan, Wan, Jiang, and Wang}]{kong2024platolmteachingllmsmultiround}
Chuyi Kong, Yaxin Fan, Xiang Wan, Feng Jiang, and Benyou Wang. 2024.
\newblock \href {https://arxiv.org/abs/2308.11534} {Platolm: Teaching llms in multi-round dialogue via a user simulator}.
\newblock \emph{Preprint}, arXiv:2308.11534.

\bibitem[{Kwon et~al.(2023)Kwon, Li, Zhuang, Sheng, Zheng, Yu, Gonzalez, Zhang, and Stoica}]{kwon2023efficient}
Woosuk Kwon, Zhuohan Li, Siyuan Zhuang, Ying Sheng, Lianmin Zheng, Cody~Hao Yu, Joseph~E. Gonzalez, Hao Zhang, and Ion Stoica. 2023.
\newblock \href {https://arxiv.org/abs/2309.06180} {Efficient memory management for large language model serving with pagedattention}.
\newblock \emph{Preprint}, arXiv:2309.06180.

\bibitem[{Li et~al.(2023)Li, Zhang, Dubois, Taori, Gulrajani, Guestrin, Liang, and Hashimoto}]{alpaca_eval}
Xuechen Li, Tianyi Zhang, Yann Dubois, Rohan Taori, Ishaan Gulrajani, Carlos Guestrin, Percy Liang, and Tatsunori~B. Hashimoto. 2023.
\newblock Alpacaeval: An automatic evaluator of instruction-following models.
\newblock \url{https://github.com/tatsu-lab/alpaca_eval}.

\bibitem[{Meng et~al.(2024)Meng, Xia, and Chen}]{meng2024simposimplepreferenceoptimization}
Yu~Meng, Mengzhou Xia, and Danqi Chen. 2024.
\newblock \href {https://arxiv.org/abs/2405.14734} {Simpo: Simple preference optimization with a reference-free reward}.
\newblock \emph{Preprint}, arXiv:2405.14734.

\bibitem[{Oliver(1977)}]{Oliver1977}
Richard~L. Oliver. 1977.
\newblock \href {https://doi.org/10.1037/0021-9010.62.4.480} {Effect of expectation and disconfirmation on postexposure product evaluations: An alternative interpretation}.
\newblock \emph{Journal of Applied Psychology}, 62(4):480--486.

\bibitem[{Oliver(1980)}]{Oliver1980}
Richard~L. Oliver. 1980.
\newblock \href {https://doi.org/10.2307/3150499} {A cognitive model of the antecedents and consequences of satisfaction decisions}.
\newblock \emph{Journal of Marketing Research}, 17(4):460--469.

\bibitem[{OpenAI et~al.(2024)OpenAI, Achiam, and et~al.}]{openai2024gpt4technicalreport}
OpenAI, Josh Achiam, and et~al. 2024.
\newblock \href {https://arxiv.org/abs/2303.08774} {Gpt-4 technical report}.
\newblock \emph{Preprint}, arXiv:2303.08774.

\bibitem[{Ouyang et~al.(2022)Ouyang, Wu, Jiang, Almeida, Wainwright, Mishkin, Zhang, Agarwal, Slama, Ray et~al.}]{ouyang2022training}
Long Ouyang, Jeffrey Wu, Xu~Jiang, Diogo Almeida, Carroll Wainwright, Pamela Mishkin, Chong Zhang, Sandhini Agarwal, Katarina Slama, Alex Ray, et~al. 2022.
\newblock Training language models to follow instructions with human feedback.
\newblock \emph{Advances in neural information processing systems}, 35:27730--27744.

\bibitem[{Rafailov et~al.(2024)Rafailov, Sharma, Mitchell, Ermon, Manning, and Finn}]{rafailov2024directpreferenceoptimizationlanguage}
Rafael Rafailov, Archit Sharma, Eric Mitchell, Stefano Ermon, Christopher~D. Manning, and Chelsea Finn. 2024.
\newblock \href {https://arxiv.org/abs/2305.18290} {Direct preference optimization: Your language model is secretly a reward model}.
\newblock \emph{Preprint}, arXiv:2305.18290.

\bibitem[{Reimers and Gurevych(2019)}]{reimers2019sentencebertsentenceembeddingsusing}
Nils Reimers and Iryna Gurevych. 2019.
\newblock \href {https://arxiv.org/abs/1908.10084} {Sentence-bert: Sentence embeddings using siamese bert-networks}.
\newblock \emph{Preprint}, arXiv:1908.10084.

\bibitem[{Schulman et~al.(2017)Schulman, Wolski, Dhariwal, Radford, and Klimov}]{schulman2017proximalpolicyoptimizationalgorithms}
John Schulman, Filip Wolski, Prafulla Dhariwal, Alec Radford, and Oleg Klimov. 2017.
\newblock \href {https://arxiv.org/abs/1707.06347} {Proximal policy optimization algorithms}.
\newblock \emph{Preprint}, arXiv:1707.06347.

\bibitem[{Sun et~al.(2024)Sun, Liu, Zhou, Huang, Song, Zhao, Zhang, Zhang, and Gai}]{sun2024parrotenhancingmultiturninstruction}
Yuchong Sun, Che Liu, Kun Zhou, Jinwen Huang, Ruihua Song, Wayne~Xin Zhao, Fuzheng Zhang, Di~Zhang, and Kun Gai. 2024.
\newblock \href {https://arxiv.org/abs/2310.07301} {Parrot: Enhancing multi-turn instruction following for large language models}.
\newblock \emph{Preprint}, arXiv:2310.07301.

\bibitem[{Ulmer et~al.(2024)Ulmer, Mansimov, Lin, Sun, Gao, and Zhang}]{ulmer2024bootstrappingllmbasedtaskorienteddialogue}
Dennis Ulmer, Elman Mansimov, Kaixiang Lin, Justin Sun, Xibin Gao, and Yi~Zhang. 2024.
\newblock \href {https://arxiv.org/abs/2401.05033} {Bootstrapping llm-based task-oriented dialogue agents via self-talk}.
\newblock \emph{Preprint}, arXiv:2401.05033.

\bibitem[{Wang et~al.(2024{\natexlab{a}})Wang, Ma, Feng, Zhang, Yang, Zhang, Chen, Tang, Chen, Lin, Zhao, Wei, and Wen}]{Wang_2024}
Lei Wang, Chen Ma, Xueyang Feng, Zeyu Zhang, Hao Yang, Jingsen Zhang, Zhiyuan Chen, Jiakai Tang, Xu~Chen, Yankai Lin, Wayne~Xin Zhao, Zhewei Wei, and Jirong Wen. 2024{\natexlab{a}}.
\newblock \href {https://doi.org/10.1007/s11704-024-40231-1} {A survey on large language model based autonomous agents}.
\newblock \emph{Frontiers of Computer Science}, 18(6).

\bibitem[{Wang et~al.(2024{\natexlab{b}})Wang, Zhang, Yang, Chen, Tang, Zhang, Chen, Lin, Song, Zhao, Xu, Dou, Wang, and Wen}]{wang2024userbehaviorsimulationlarge}
Lei Wang, Jingsen Zhang, Hao Yang, Zhiyuan Chen, Jiakai Tang, Zeyu Zhang, Xu~Chen, Yankai Lin, Ruihua Song, Wayne~Xin Zhao, Jun Xu, Zhicheng Dou, Jun Wang, and Ji-Rong Wen. 2024{\natexlab{b}}.
\newblock \href {https://arxiv.org/abs/2306.02552} {User behavior simulation with large language model based agents}.
\newblock \emph{Preprint}, arXiv:2306.02552.

\bibitem[{Wang et~al.(2023)Wang, Tang, Zhao, Wang, and Wen}]{iEvalLM}
Xiaolei Wang, Xinyu Tang, Xin Zhao, Jingyuan Wang, and Ji-Rong Wen. 2023.
\newblock \href {https://doi.org/10.18653/v1/2023.emnlp-main.621} {Rethinking the evaluation for conversational recommendation in the era of large language models}.
\newblock In \emph{Proceedings of the 2023 Conference on Empirical Methods in Natural Language Processing}. Association for Computational Linguistics.

\bibitem[{Wang et~al.(2022)Wang, Zhou, Wen, and Zhao}]{Wang_2022}
Xiaolei Wang, Kun Zhou, Ji-Rong Wen, and Wayne~Xin Zhao. 2022.
\newblock \href {https://doi.org/10.1145/3534678.3539382} {Towards unified conversational recommender systems via knowledge-enhanced prompt learning}.
\newblock In \emph{Proceedings of the 28th ACM SIGKDD Conference on Knowledge Discovery and Data Mining}, KDD ’22, page 1929–1937. ACM.

\bibitem[{Wei et~al.(2023)Wei, Wang, Schuurmans, Bosma, Ichter, Xia, Chi, Le, and Zhou}]{wei2023chainofthoughtpromptingelicitsreasoning}
Jason Wei, Xuezhi Wang, Dale Schuurmans, Maarten Bosma, Brian Ichter, Fei Xia, Ed~Chi, Quoc Le, and Denny Zhou. 2023.
\newblock \href {https://arxiv.org/abs/2201.11903} {Chain-of-thought prompting elicits reasoning in large language models}.
\newblock \emph{Preprint}, arXiv:2201.11903.

\bibitem[{Wells et~al.(1971)Wells, Tigert, and Activities}]{wells1971opinions}
William~D Wells, Douglas~J Tigert, and Interests Activities. 1971.
\newblock Opinions.
\newblock \emph{Journal of advertising research}, 11(4):27--35.

\bibitem[{Xie et~al.(2024)Xie, Goyal, Zheng, Kan, Lillicrap, Kawaguchi, and Shieh}]{xie2024montecarlotreesearch}
Yuxi Xie, Anirudh Goyal, Wenyue Zheng, Min-Yen Kan, Timothy~P. Lillicrap, Kenji Kawaguchi, and Michael Shieh. 2024.
\newblock \href {https://arxiv.org/abs/2405.00451} {Monte carlo tree search boosts reasoning via iterative preference learning}.
\newblock \emph{Preprint}, arXiv:2405.00451.

\bibitem[{Yao et~al.(2023)Yao, Zhao, Yu, Du, Shafran, Narasimhan, and Cao}]{yao2023react}
Shunyu Yao, Jeffrey Zhao, Dian Yu, Nan Du, Izhak Shafran, Karthik Narasimhan, and Yuan Cao. 2023.
\newblock {ReAct}: Synergizing reasoning and acting in language models.
\newblock In \emph{International Conference on Learning Representations (ICLR)}.

\bibitem[{Zhang et~al.(2024)Zhang, Xie, Lyu, Xin, Ren, Liang, Zhang, Kang, de~Rijke, and Ren}]{zhang2024towards}
Xiaoyu Zhang, Ruobing Xie, Yougang Lyu, Xin Xin, Pengjie Ren, Mingfei Liang, Bo~Zhang, Zhanhui Kang, Maarten de~Rijke, and Zhaochun Ren. 2024.
\newblock Towards empathetic conversational recommender systems.
\newblock In \emph{Proceedings of the 18th ACM Conference on Recommender Systems}, pages 84--93.

\bibitem[{Zheng et~al.(2024)Zheng, Zhang, Zhang, Ye, Luo, Feng, and Ma}]{zheng2024llamafactory}
Yaowei Zheng, Richong Zhang, Junhao Zhang, Yanhan Ye, Zheyan Luo, Zhangchi Feng, and Yongqiang Ma. 2024.
\newblock \href {http://arxiv.org/abs/2403.13372} {Llamafactory: Unified efficient fine-tuning of 100+ language models}.
\newblock In \emph{Proceedings of the 62nd Annual Meeting of the Association for Computational Linguistics (Volume 3: System Demonstrations)}, Bangkok, Thailand. Association for Computational Linguistics.

\bibitem[{Zhu et~al.(2024)Zhu, Huang, and Sang}]{zhu2024llmbasedcontrollablescalablehumaninvolved}
Lixi Zhu, Xiaowen Huang, and Jitao Sang. 2024.
\newblock \href {https://arxiv.org/abs/2405.08035} {A llm-based controllable, scalable, human-involved user simulator framework for conversational recommender systems}.
\newblock \emph{Preprint}, arXiv:2405.08035.

\end{thebibliography}

\newpage
\appendix
\newpage
\appendix


\newpage
\clearpage
\section{Related Work}
\paragraph{Conversational Recommendation Systems.} A CRS aims to engage users through natural language interaction, iteratively eliciting their preferences and providing personalized recommendations~\citep{zhang2024towards}. Research on CRS can be divided into two categories: attribute-based CRSs and generation-based CRSs~\citep{CRS_Survey_Jannach}. While attribute-based CRSs rely on pre-defined templates, generation-based CRSs~\citep{Wang_2022} enable more flexible interactions but are constrained by the limitations of traditional language models. 
In recent years, leveraging the powerful language capabilities and tool utilization of LLMs, researchers have begun developing CRAs, offering transformative solutions for conversational recommendation. ZSCRS~\citep{he2023zscrs} conducts an initial exploration of using LLMs directly as conversational recommenders. ChatRec~\citep{gao2023chatrec} and InteRecAgent~\citep{huang2023interecagent} integrate traditional recommendation models with LLMs, effectively enhancing the interactivity of the recommendation system. 
To further enhance dialogue flow control in CRS, MACRS~\citep{fang2024multi} introduces a multi-agent framework to enable long-term strategic planning. Despite the extensive exploration of LLM-based frameworks for CRAs, the increasing inference costs of complex agent frameworks and the diminishing returns on performance gains have significantly limited their practicality in real-world scenarios. Hence, there is an urgent need for an alignment method that fine-tunes LLMs for CRA tasks, enabling them to better guide users and effectively enhance user satisfaction.

\paragraph{LLM Alignment.}
The objective of LLMs is to predict the next token in internet-scale corpora; however, this differs from the goal of "helpfully and safely following the user's instructions"~\cite{ouyang2022training}. Therefore, it is necessary to align LLMs with human preferences to ensure the generation of safe, unbiased, and appropriate text~\citep{schulman2017proximalpolicyoptimizationalgorithms, rafailov2024directpreferenceoptimizationlanguage, ethayarajh2024ktomodelalignmentprospect, meng2024simposimplepreferenceoptimization}.
In this paper, we focus on the problem of LLM alignment in multi-turn conversational recommendation (MTPO).
Currently, most existing methods~\citep{sun2024parrotenhancingmultiturninstruction, ulmer2024bootstrappingllmbasedtaskorienteddialogue, kong2024platolmteachingllmsmultiround} simply treat each turn equally, failing to capture turn-level preference relationships. Another class of methods~\citep{jin2024persuading, xie2024montecarlotreesearch} employs tree-based simulation to infer \textbf{turn-level} preference relations. Specifically, these methods generate multiple candidate outputs at each intermediate turn, expand them into different dialogue subpaths, and simulate complete dialogues to obtain final rewards, thereby estimating the rewards for intermediate turns. However, these methods introduce additional sampling costs and struggle to establish high-quality preference relationships. Therefore, developing a method to construct high-quality turn-level preference relationships without additional sampling and evaluation is critical.

\begin{table*}[t]
\centering
\renewcommand{\arraystretch}{1}
\begin{tabular}{l *{9}{c}} 
\toprule
\multirow{2}{*}{\textbf{Method}} 
& \multicolumn{3}{c}{\textbf{ChatRec}} 
& \multicolumn{3}{c}{\textbf{ReAct}}
& \multicolumn{3}{c}{\textbf{ActCRS}} \\
\cmidrule(lr){2-4} \cmidrule(lr){5-7} \cmidrule(lr){8-10} 
& \textbf{SR} & \textbf{R} & \textbf{WR}
& \textbf{SR} & \textbf{R} & \textbf{WR}
& \textbf{SR} & \textbf{R} & \textbf{WR} \\
\midrule
GPT-4o mini
& 0.46 & 0.47 & 0.50
& 0.52 & 0.56 & 0.50
& 0.53 & 0.56 & 0.50 \\
\midrule
Llama-3.1
& 0.42 & 0.47 & 0.11 
& 0.36 & 0.54 & 0.31
& 0.34 & 0.55 & 0.28 \\
\blue{\textit{+SGPT(Ours)}} 
& \blue{\underline{0.47}} & \blue{\underline{0.53}} & \blue{\underline{0.39}} 
& \blue{\textbf{0.54}} & \blue{\textbf{0.61}} & \blue{\underline{0.41}}
& \blue{\underline{0.54}} & \blue{\underline{0.57}} & \blue{\underline{0.48}} 
\\
\blue{\textit{+ECPO(Ours)}} 
& \blue{\textbf{0.50}} & \blue{\textbf{0.54}} & \blue{\textbf{0.41}} 
& \blue{\textbf{0.54}} & \blue{\underline{0.60}} & \blue{\textbf{0.49}} 
& \blue{\textbf{0.56}} & \blue{\textbf{0.58}} & \blue{\textbf{0.57}} \\
\bottomrule
\end{tabular}
\caption{Effectiveness on different CRAs.}
\label{tab:different CRA}
\end{table*}

\section{Experimental Details}
\label{app: exp_details}
\subsection{Simulator Details}
Traditional CRS evaluation methods struggle to assess dynamic CRA tasks~\citep{Afzali_2023}. As discussed in Section~\ref{sec: ailo}, we follow and extend iEvalLM~\citep{iEvalLM} by introducing AILO for our evaluations. Our experiments utilize the Amazon-Game, Amazon-Book\footnote{\url{https://github.com/hyp1231/AmazonReviews2023}}, and Yelp\footnote{\url{https://www.yelp.com/dataset}} datasets, constructing 100 user personas for each dataset. We use GPT-4o mini as the backbone model of the AILO user simulator.

\subsection{CRA Task Details}
\paragraph{Task Construction.} To generate high-quality conversational recommendation tasks, we extract positively rated items from each user's interaction history and designate them as ground-truth items for recommendation tasks. As a result, each user is assigned approximately 30 conversational recommendation tasks, yielding 3,000 simulation tasks per dataset (100 × 30). Additionally, we use all-MiniLM-L6-v2~\citep{reimers2019sentencebertsentenceembeddingsusing} to embed items users have interacted with, constructing an external database $I$.

\paragraph{Task Execution.} We model conversational recommendation as an agent-based task. As shown in Figure~\ref{fig:ailo}, during task execution, the CRA engages in multi-turn interactions with AILO via natural language and can optionally query the external retriever to obtain real item information from the database \( I \) for making recommendations.
The primary objective of the CRA is to recommend the ground truth item using natural language. Depending on the agent framework, the retrieval query may vary: (1) some frameworks directly concatenate the dialogue history as input (ChatRec~\citep{gao2023chatrec}), while (2) others generate the query using an LLM. To foster further research, we have open-sourced the automated task generation process.

\subsection{Implementation Details}
\paragraph{Inference Details.} In the main experiments, each CRA task consists of a maximum of 5 interaction turns, and each retiever will returns the top 5 most relevant items with the retrieval query. Following \citet{yao2023react, fang2024multi}, we sample 100 tasks for each dataset for testing. In all LLM inference processes, we set the temperature parameter of all models to 0.0 to eliminate randomness in local models. Additionally, we use the vllm~\citep{kwon2023efficient} framework to accelerate all inference processes.

\paragraph{Training Details.}~During training, we use the Llama-factory~\citep{zheng2024llamafactory} for LoRA~\citep{hu2021lora} training. In our main experiments, we set the rewriting threshold to $\lambda = 4.0$. We randomly sample 1000 simulation tasks to construct the dataset $D_\text{sft}$, generating approximately 2000 input-output pairs, and randomly sample an additional 500 tasks for constructing dataset $D_\text{pre}$.  For the stage of SGPT, we use a learning rate of $5e-05$ and a batch size of 8. For the stage of ECPO, we search two learning rates: $\{1e-06, 5e-07\}$, with a batch size of 32.

\subsection{Human Evaluation Details}
We conduct three sets of human evaluation experiments: (1) empirical verification of data before and after rewriting, (2) comparison of the reliability between AILO and iEval, and (3) evaluation of ECPO against expert models.  In each experiment, we employ two annotators with an average education level of a bachelor's degree. To ensure fairness, we randomly shuffle and anonymize the data before annotation. This blind evaluation setup minimizes potential biases and improves the reliability of our results.

\subsection{Evaluation Metrics.}
\label{app: metrics}
We evaluate CRAs across two dimensions.
\begin{itemize}
    \item \textbf{Recommendation Metrics}: Following iEvalLM~\citep{iEvalLM}, we report \textbf{SR} (success rate of recommending the ground-truth item) and \textbf{R} (rate of retrieving the ground-truth item from the item database).
    \item \textbf{Dialogue Metrics}: We use \textbf{WR} (win rate~\citep{alpaca_eval}) to assess dialogue quality, focusing on flexibility, coherence, and user guidance ability. Specifically, we use GPT-4o~\citep{openai2024gpt4technicalreport} to evaluate dialogue quality by comparing the target model's responses with those of the expert CRA. We select GPT-based ActCRS as the expert CRA because it demonstrated the best interactive performance in our preliminary validation experiments. The win rate is then calculated based on these comparisons. The evaluation prompt design is shown in Figure~\ref{tab:prompt_traj_eval}. To mitigate potential positional bias, we conduct evaluations by swapping the positions of \textit{Traj\_a} and \textit{Traj\_b} twice and averaging the scores to obtain the final result.
\end{itemize}

\section{Further Analysis}

\subsection{Effectiveness of ECPO on Different CRAs}
\label{sec: dcra}
In this section, we perform SGPT+ECPO fine-tuning on different CRA frameworks (ChatRec, ReAct, ActCRS) using the Amazon-Book dataset. To ensure clarity, within each CRA framework, we consistently use the GPT-based CRA as the comparison baseline when calculating win rates.
 As shown in~\ref{tab:different CRA}, we find that, overall, all frameworks exhibit significant improvements over the original CRA after fine-tuning, with their performance approaching or even surpassing that of GPT.
This superior performance confirms the applicability of our method across various CRAs.

\subsection{Empirical Evaluation of Rewritten vs. Unsatisfactory Responses}
\label{sec: emofrewrite}

\begin{table}[t]
    \centering
    \begin{tabular}{lccc}
        \toprule
        \textbf{Evaluator} & \textbf{Win} & \textbf{Tie} & \textbf{Lose} \\
        \midrule
        GPT-4o  & 0.64 & 0.34 & 0.02 \\
        Human  & 0.80 & 0.20 & 0.04 \\
        \bottomrule
    \end{tabular}
    \caption{Win rate of Rewritten response vs. Unsatisfactory responses.}
    \label{tab:gpt4_human_comparison}
\end{table}

In this section, following~\citet{wang2024userbehaviorsimulationlarge}, we compare user satisfaction before and after rewriting the responses using both GPT-4o and human annotators. As shown in Table~\ref{tab:gpt4_human_comparison}, evaluations from both GPT-4o and human annotators indicate that rewritten responses are \textbf{predominantly superior} to the unsatisfactory ones, with only a few instances where they perform slightly worse. Additionally, human annotators tend to assign more wins, whereas GPT-4o produces more ties. We hypothesize that this discrepancy arises because humans are more attuned to subtle variations in dialogue style.  
These empirical findings confirm that rewritten responses are statistically more likely to outperform the original ones. Furthermore, the exceptional performance of ECPO in the main experiments further substantiates this claim.

\subsection{Case Studies}
\label{sec: case}
In this section, we conduct a case study to analyze how ECPO modifies dialogue style and better aligns with user expectations. We present two dialogue examples: Llama-3.1+SGPT (Figure \ref{tab:case_sgpt}) and Llama-3.1+ECPO (Figure \ref{tab:case_ecpo}).
We find that Llama-3.1+SGPT exhibits lower proactivity and is prone to falling into error loops. This is mainly evident when the user's needs are unclear, and the system fails to effectively guide the conversation, resulting in repeated questioning of the same issues or providing irrelevant recommendations. In contrast, Llama-3.1+ECPO demonstrates stronger proactivity and flexibility. Through precise demand guidance and strategy adjustments, the model actively identifies user needs and makes more appropriate recommendations, thereby significantly improving the fluidity of the conversation and user satisfaction.

\section{Prompts}
In this section, we present the core prompt design for forward expectation confirmation 
($I_{\text{ect}}$, Figure~\ref{tab:prompt_ec}) and backward expectation derivation ($I_{\text{bed}}$, Figure~\ref{tab:bed}). 

\begin{figure*}[h]
  \centering
  \begin{tcolorbox}[
      width=\textwidth,           
      enhanced,                   
colframe=magenta!40!black,
    colback=magenta!5!white,
    colbacktitle=magenta!30!black,
      coltitle=white,             
      title=\Large \textbf{Case Study: Llama-3.1+SGPT},   
      fonttitle=\bfseries,
      arc=4mm,                    
      boxsep=5pt,                 
      top=5pt, bottom=5pt,        
  ]
  \raggedright
    \textbf{User:} I'm interested in a book that explores themes of friendship and adjustment in a new environment! \\ [2ex]
    \textbf{Assistant:} What age group or genre are you looking for in the book? For example, are you interested in young adult, adult fiction, or perhaps a specific type of story like coming-of-age or contemporary? \\[2ex]
    \textbf{User:} I'm particularly interested in children's literature, especially stories that resonate with younger readers. A coming-of-age theme would be delightful, as it often captures the essence of friendship and the challenges of adapting to new surroundings! \\ [2ex]
    \textbf{Assistant:} "It's OK to Need a Friend (Little Brown Bear)" teaches the importance of empathy and kindness in friendships, making it a poignant read for younger readers. Alternatively, "Side By Side" celebrates friendship through the journey of a little mouse seeking a best friend in Wintermouse Wood. Would you like more details about either of these books? \textcolor{blue}{(The theme of children's growth is captured, but it deviates from literary novels.)}\\[2ex]
    \textbf{User:} These books sound nice, but I'm looking for something that specifically explores the immigrant experience and the challenges of adapting to a new home. Are there any other options that fit that theme? \\[2ex]
    \textbf{Assistant:} "What If All the Kids Are White?: Anti-bias Multicultural Education With Young Children And Families" addresses the challenges of adapting to a new environment through the lens of racial and cultural diversity, providing strategies for teaching children about different cultures and identities. While it may not be a traditional coming-of-age story, it resonates with the themes of adjustment and understanding in a new home. Would you like more information about this book or perhaps explore other options? \textcolor{blue}{(CRA again ignores users' desire for literary fiction.)}\\[2ex]
    \textbf{User:} This book touches on important themes, but I'm really looking for something that specifically highlights the immigrant experience and the challenges of adapting to a new home. Are there any other options that fit that theme? \\[2ex]
    \textbf{Assistant:} "What If All the Kids Are White?: Anti-bias Multicultural Education With Young Children And Families" addresses the challenges of adapting to a new environment through the lens of racial and cultural diversity, providing strategies for teaching children about different cultures and identities. While it may not be a traditional coming-of-age story, it resonates with the themes of adjustment and understanding in a new home. Unfortunately, it seems there are limited options specifically focused on the immigrant experience in children's literature. Would you like to explore more general themes or perhaps look into adult literature that addresses this topic? \textcolor{blue}{(Error Loop.)}\\[2ex]
    \textbf{User:} This recommendation is interesting, but it doesn't quite capture the immigrant experience I'm seeking. Are there any other children's books that focus more directly on the challenges faced by immigrant children? \\[2ex]
    \textbf{Assistant:} "A Kid's Guide to African American History: More than 70 Activities" offers engaging activities that teach children about African American history and culture, which may provide insights into the experiences of immigrant communities. ... \textcolor{blue}{(Error Loop.)}\\
\end{tcolorbox}
\caption{SGPT-aligned CRA is relatively rigid and lacks proactivity, making it susceptible to falling into error loops.}
    \label{tab:case_sgpt}
\end{figure*}

\begin{figure*}[h]
  \centering
  \begin{tcolorbox}[
      width=\textwidth,           
      enhanced,                   
       colframe=blue!50!black,     
      colback=blue!10!white,       
      colbacktitle=blue!60!black, 
      coltitle=white,             
      title=\Large \textbf{Case Study: Llama-3.1+ECPO},   
      fonttitle=\bfseries,
      arc=4mm,                    
      boxsep=5pt,                 
      top=5pt, bottom=5pt,        
  ]
\textbf{User:} I'm looking for a book that captures the essence of growing up and finding connection in a new place! \\ [2ex]
    \textbf{Assistant:} What age group are you interested in for the book? Are you looking for a specific genre, like young adult, coming-of-age, or perhaps a more general fiction? Additionally, do you prefer a more serious tone or something lighter and humorous? \textcolor{blue}{(CRA Ask more specific clarifying questions that were easy to answer without placing additional burden on the user.)}\\[2ex]
    \textbf{User:} I'm open to a range of age groups, but I think something that appeals to both young adults and perhaps even a broader audience would be ideal! I love stories that are rich in emotional depth, so a coming-of-age narrative would be fantastic. While I appreciate serious themes, I also enjoy a lighter touch or humor woven into the story to balance the emotional journey! \\ [2ex]
    \textbf{Assistant:} "The Perks of Being a Wallflower" is a poignant coming-of-age story that beautifully captures the essence of growing up and finding connection in a new place. The narrative follows Charlie, a shy teenager navigating high school and friendships while grappling with his own emotional struggles. The book balances serious themes with lighter moments, making it relatable and engaging for both young adults and a broader audience. Would you like to know more about this book or explore other recommendations? \textcolor{blue}{(CRA correctly grasp that users are interested in youth, growth and emotional novels.)}\\[2ex]
    \textbf{User:} This sounds interesting, but I'm particularly looking for a story that focuses more on the immigrant experience and the sense of community that comes with it. Are there other recommendations that fit that theme? \textcolor{blue}{(The dialogue does not deviate from the theme of novel.)}\\[2ex]
    \textbf{Assistant:} Are you interested in stories that specifically focus on the challenges faced by immigrants, or would you like to explore themes of cultural identity and community as well? Additionally, do you prefer a more serious tone, or are you open to lighter narratives that still convey the essence of the immigrant experience?  \textcolor{blue}{(Further confirmation of details.)}\\[2ex]
    \textbf{User:} I'm definitely interested in stories that delve into the challenges faced by immigrants, as well as those that explore themes of cultural identity and the sense of community that develops in new environments. I appreciate narratives that can convey the emotional weight of these experiences, but I also enjoy when there's a lighter touch or humor woven in to create a more balanced reading experience. It’s fascinating how these stories can reflect both the struggles and the joys of finding one’s place in a new world!\\[2ex]
    \textbf{Assistant:} \textcolor{blue}{"Rosa's Song" is a poignant picture book that beautifully captures the immigrant experience through the eyes of Jae, a young Korean boy adjusting to life in a new apartment building filled with other newcomers.} The story highlights ... Would you like to explore more recommendations that focus on the immigrant experience and cultural identity, or would you like to delve deeper into this book? \textcolor{blue}{(CRA correctly recommends target items.)}\\[2ex]
    \textbf{User:} This sounds lovely and captures the essence of the immigrant experience beautifully! ... \\[2ex]
\end{tcolorbox}
    \caption{ECPO-aligned CRA exhibits greater proactivity and a more flexible interaction style in dialogue, actively guiding users and successfully providing recommendations.}
    \label{tab:case_ecpo}
\end{figure*}

\begin{figure*}[h]
  \centering
  \begin{tcolorbox}[
      width=\textwidth,           
      enhanced,                   
    colframe=brown!30!white,     
    colback=brown!10!white,      
    colbacktitle=brown!40!black, 
      coltitle=white,             
      title=\Large \textbf{Prompt Design for Dialogue Evaluation},   
      fonttitle=\bfseries,
      arc=4mm,                    
      boxsep=5pt,                 
      top=5pt, bottom=5pt,        
  ]
    You are provided with two dialogue trajectories for comparison. Evaluate each dialogue system using the following criteria: \\[3pt]
    \textbf{1. Flexibility:} How well does the system adapt to changes in user requests or shifts in conversation flow? \\[2pt]
    \quad \quad 1: A is more flexible. \\
    \quad \quad -1: B is more flexible. \\
    \quad \quad 0: Both are equally flexible. \\[3pt]
    \textbf{2. Coherence:} How consistent and fluid is the dialogue? Does the system remember context and respond appropriately to the user's input? \\[2pt]
    \quad \quad 1: A is more coherent. \\
    \quad \quad -1: B is more coherent. \\
    \quad \quad 0: Both are equally coherent. \\[3pt]
    \textbf{3. User Guidance:} How well does the system guide the user, clarify requests, or steer the conversation in a productive direction? \\[2pt]
    \quad \quad 1: A provides better guidance. \\
    \quad \quad -1: B provides better guidance. \\
    \quad \quad 0: Both provide similar levels of guidance. \\[3pt]
    \textbf{4. Overall Evaluation:} Based on the above three indicators, determine which trajectory is better. \\[3pt]
    \textbf{Trajectory A:} \{Traj\_a\} \\[2pt]
    \textbf{Trajectory B:} \{Traj\_b\} \\[3pt]
    Please provide a score of \textbf{1}, \textbf{-1}, or \textbf{0} based on the comparison. After scoring, output the result in the following pure JSON format: \\[5pt]
    {\texttt{\{}} \\ 
    \quad {\texttt{"Flexibility": \{}} \\ 
    \quad \quad {\texttt{"Reason": "{reason}",}} \\ 
    \quad \quad {\texttt{"Score": -1 or 1 or 0}} \\ 
    \quad {\texttt{\},}} \\ 
    \quad {\texttt{"Coherence": \{}} \\ 
    \quad \quad {\texttt{"Reason": "{reason}",}} \\ 
    \quad \quad {\texttt{"Score": -1 or 1 or 0}} \\ 
    \quad {\texttt{\},}} \\ 
    \quad {\texttt{"User Guidance": \{}} \\ 
    \quad \quad {\texttt{"Reason": "{reason}",}} \\ 
    \quad \quad {\texttt{"Score": -1 or 1 or 0}} \\ 
    \quad {\texttt{\},}} \\ 
    \quad {\texttt{"Final Score": -1 or 1 or 0}} \\ 
    {\texttt{\}}} \\[5pt]

\end{tcolorbox}
    \caption{Prompt design for evaluating dialogue trajectories.}
    \label{tab:prompt_traj_eval}
\end{figure*}

\begin{figure*}[h]
  \centering
  \begin{tcolorbox}[
      width=\textwidth,           
      enhanced,                   
colframe=red!50!white,     
    colback=red!3!white,       
    colbacktitle=red!60!black, 
      coltitle=white,             
      title=\Large \textbf{Prompt Design for Forward Expectation Confirmation},   
      fonttitle=\bfseries,
      arc=4mm,                    
      boxsep=5pt,                 
      top=5pt, bottom=5pt,        
  ]
You are a user simulator, and your task is to evaluate the expressiveness and interaction quality of the {domain} conversational recommendation system in its last interaction.  
    Your evaluation should focus on how well the system's response supports the dialogue flow, user engagement, and natural communication. \\

    \textbf{1. Evaluation Dimensions:} \\
    \quad \quad \textbf{Flexibility}: How well does the system adapt its responses to changes in user requests or shifts in the conversation flow? \\
    \quad\quad \quad \textbf{Score Range}: 0 to -2 points \\
    \quad\quad \quad \textbf{Deductions}: \\
    \quad\quad \quad \quad -2 points: The system fails to recognize and respond to the user's change in intent or request, resulting in a rigid, non-adaptive response. \\
    \quad\quad \quad \quad -1 point: The system identifies the change in intent but responds in a delayed, overly rigid, or awkward manner. \\
    \quad\quad \quad \quad 0 points: The system fully adapts to changes in user requests, showing natural flexibility in its responses. \\
    \quad \quad \textcolor{blue}{...Descriptions of Coherence and User guidance}\\

    \textbf{2. Scoring Method:} \\
    \quad\quad 1.The initial score is \textbf{5 points} (Flexibility = 2, Coherence = 2, User Guidance = 1). \\
    \quad\quad 2.Points are deducted based on the criteria outlined above for each dimension. \\
    \quad\quad 3.Final Score = 5 - (Flexibility deductions) - (Coherence deductions) - (User Guidance deductions) \\
    \quad\quad Score Range: 0 to 5 points (higher score indicates better expressiveness and interactivity). \\

    \textbf{3. Feedback Requirement:} \\
    \quad\quad  1.Provide a reason for the score, referencing specific aspects of the system’s expressiveness (e.g., its flexibility, coherence, and user guidance). \\
    \quad\quad  2.Highlight any specific user reactions (e.g., confusion, frustration, or engagement) that support the score. \\
    \quad\quad  3.Clearly mention the specific issues that caused point deductions, such as rigid responses, logical inconsistencies, or lack of guidance. \\

    \textbf{Inputs:} \\
    \quad\quad System's Last Response: \{last\_turn\_response\} \\
    \quad\quad Dialogue History: \{Dialogue\_history\} \\
    \quad\quad Target: \{Target\_item\} \textcolor{blue}{\#In the actual implementation, the target item name in this instruction is optional because it has already been provided to the user during response generation.}\\
    
    \textbf{Output the results strictly in the following JSON format:} \\
    \quad \texttt{\{} \\
    \quad \quad \texttt{"reason": "\textless The reason for the score, referencing specific aspects of the system’s expressiveness, including its flexibility, coherence, and user guidance. Mention the specific issues that led to deductions. \textgreater"}, \\
    \quad \quad \texttt{"rating": "\textless Final rating from 0 to 5 \textgreater"} \\
    \quad \texttt{\}} \\
\end{tcolorbox}
    \caption{Prompt design for forward expectation confirmation.}
    \label{tab:prompt_ec}
\end{figure*}

\begin{figure*}[h]
  \centering
  \begin{tcolorbox}[
      width=\textwidth,           
      enhanced,                   
      colframe=teal!50!black,       
      colback=teal!10!white,        
      colbacktitle=teal!80!black,   
      coltitle=white,               
      title=\Large \textbf{Prompt Design for Backward Expectation Derivation},   
      fonttitle=\bfseries,
      arc=4mm,                    
      boxsep=5pt,                 
      top=5pt, bottom=5pt,        
  ]
  
    You are a rewrite model, and your task is to improve the system's response in a conversational recommendation agent (CRS).  
    The CRS solves the task by interleaving \textbf{"Observation"} and \textbf{"Action"} steps.  
    Observations include user requests, replies, or search results retrieved by the CRS.  
    The CRS interacts with the user and the environment by taking one of the following four actions:  
    ... \\

    \textbf{Inputs Provided:} \\
    \quad\quad Scratchpad: The agent's previous interaction history. \\
    \quad\quad Original Response: The system's original response that needs improvement. \\
    \quad\quad Feedback on Flaws: Specific feedback on identified weaknesses in \textit{Flexibility}, \textit{Coherence}, and \textit{User Guidance}. \\

    \textbf{Task:} \\
    \quad\quad Your goal is to generate a \textit{rewritten response} that specifically addresses the identified flaws in \textit{Flexibility}, \textit{Coherence}, and \textit{User Guidance}. \\

    \textbf{Rewrite Strategy:} \\
    \quad\quad Targeted Flaw Fixing: Use feedback on flexibility, coherence, and user guidance as a blueprint for improvements. \\
    \quad\quad Context-Aware Rewriting: Use the conversation history to ensure the response maintains logical flow, context relevance, and user intent alignment. \\
    \quad\quad... \textcolor{blue}{\# More detailed considerations when rewriting.} \\
    
    \textbf{Inputs:} \\
    \quad\quad Scratchpad: \{Scratchpad\} \\
    \quad\quad Original Response: \{Original\_response\} \\
    \quad\quad Feedback on Flaws: \{Feedback\_flaws\} \\
    
    \textbf{Output Format:} \\
    \quad Please output the results strictly in the following JSON format: \\
    \quad \texttt{\{} \\
    \quad \quad \texttt{"reason": "\textless Reason for refinement, referencing flexibility, coherence, and user guidance improvements.\textgreater"}, \\
    \quad \quad \texttt{"refinement": "\textless Rewritten response (Ask[Question], Recommend[Answer], Response[Content] or Search[Keyword])\textgreater"} \\
    \quad \texttt{\}} \\
\end{tcolorbox}
    \caption{Prompt design for backward expectation derivation.}
    \label{tab:bed}
\end{figure*}

\end{document}